\documentstyle[epic,eepic,rotate,pnhsp,namedreferences]{kluwer}

\begin{opening}
\title{Proof Nets and the Complexity of 
      Processing Center-Embedded Constructions}
\author{Mark \surname{Johnson}\thanks{
This research was performed while on a sabbatical visit to the Rank
Xerox Research Centre in Grenoble.  I would like to thank Brown and RXRC
for making this possible, my colleagues who encouraged me to
learn about Lambek Categorial Grammar and Linear Logic, and Ted
Gibson for helpful information about human performance on center-embedded
constructions.  In addition, I would like to thank the
JoLLI reviewers for their helpful and insightful comments on the
first draft of this paper.}}
\institute{Brown University}
\end{opening}

\begin{ao} \ \\
Department of Cognitive and Linguistic Sciences, Box 1978 \\
Brown University \\
Providence, RI 02912, USA \\
Email: {\sf\makeatletter mj@lx.cog.brown.edu}
\end{ao}

\newcommand{\namecite}[1]{\citeauthor{#1} (\citeyear{#1})}
\newcommand{\SetFigFont}[5]{\small }

\begin{document}  

\maketitle
\begin{abstract}
This paper shows how proof nets can be used to formalize the notion of
``incomplete dependency'' used in psycholinguistic theories of the
unacceptability of center-embedded constructions.  Such theories of
human language processing can usually be restated in terms of
geometrical constraints on proof nets.  The paper ends with a
discussion of the relationship between these constraints and
incremental semantic interpretation.
\end{abstract}
\section{Introduction}
The distinction between competence and performance popularized by
\namecite{Chomsky65} is fundamental to modern linguistics.
{\em Competence} is the ``knowledge'' of a human language possessed by
its users (e.g., as formalized in a grammar), while {\em performance}
refers to how this knowledge is used to produce and comprehend
utterances.  To date, most applications of logic to natural language
have concentrated on competence grammar.  This paper suggests that
techniques from logic such as proof nets \cite{Girard95} might also be
useful for describing the difficulty of processing center-embedded
constructions, which is an aspect of human linguistic performance.

It is well-known that center embedded constructions such as
(\ref{e:ce}) are more difficult to comprehend than corresponding
right embedded constructions (\ref{e:re}).  (The prefix `\bad' indicates
the example is difficult or impossible to comprehend).
\eenumsentence{ 
\item \label{e:re}
The patient was cured by the drug [ that was administered by
the intern [ that was supervised by the nurse ] ]. 
\item \label{e:ce}
\bad The drug [ that the intern [ that the nurse supervised~] administered~]
cured the patient. }
As \namecite{Chomsky65} notes, the difficulty of comprehending
(\ref{e:ce}) does not seem to be due to any single construction in
(\ref{e:ce}), but rather to the way those structures are
configured.  He argues that (\ref{e:ce}) should be treated as
grammatical (i.e., possessing a well-formed syntactic analysis),
but that its incomprehensibility arises from an inability
of the human sentence processing mechanism to find that analysis.

\namecite{Chomsky63} propose that  short-term memory overload
causes this incompleteness.  
The strictly right embedded constructions of the
kind (\ref{e:re}) can be generated by a right-linear grammar, and
hence can be recognized by a finite state machine.  In contrast, a simple
application of the pumping lemma for regular sets shows that 
center embedded constructions of the kind (\ref{e:ce}) cannot
be recognized by a finite state machine.

However, it seems that all such comprehension difficulties cannot
completely be explained in terms of the memory requirements of a
parsing automaton.  As \namecite{Chomsky65} and others note, the
particular constructions and their arrangement affect
comprehensibility; thus the processing difficulty of (\ref{e:ce})
relative to (\ref{e:re}) may be due in part to the fact that in
(\ref{e:ce}) the relative clauses involve object extraction, while in
(\ref{e:re}) the relative clauses involve subject extraction.

The following example, from \namecite{Gibson96}, shows that depth of
center embedding alone cannot account for all such processing
difficulty.  (\ref{e:scrc}) and (\ref{e:rcsc}) are both doubly
embedded structures consisting of a relative clause (\RC) and a
sentential complement (\SC), differing only in the order of embedding,
and most phrase-structure parsing automata require the same amount of
working memory to accept them both.  However (\ref{e:scrc}) is easier
to comprehend than (\ref{e:rcsc}).
\eenumsentence{ \label{e:gibson}
\item \label{e:scrc}
The chance \const{SC}{ that the nurse \const{RC}{ who the doctor
supervised } lost the reports } bothered the intern.
\item \label{e:rcsc}
\bad The intern \const{RC}{ who the chance \const{SC}{ that the doctor 
lost the reports~} bothered } supervised the nurse. }
The literature contains a number of proposals accounting for a variety
of comprehensibility differences \cite{Gibson96,Lewis96,Stabler94}.
These theories of processing overload differ in the predictions that
they make, this paper does not attempt to choose between them, or to
develop an alternative account.  (Indeed, there seems to be
considerable uncertainty among theorists as to exactly which sentences
cause processing overload).

However, all these proposals share the common feature that they posit
that the human sentence processing mechanism is sensitive to the
number and kind of ``incomplete dependencies'' encountered as the
sentence is processed from left to right.  The precise configurations
of dependencies which purportedly lead to processing overload differ
from theory to theory.  Unfortunately, the central notion of
incomplete dependency, and its relationship both to grammar and to
parsing mechanisms, is usually only informally explained in these
proposals.

This paper proposes that resource logics provide the appropriate tools
to formalize and investigate such incomplete dependencies.  The next
section introduces resource logics and proof nets, and the following
section uses these to express an account of the processing complexity
of the examples just discussed.  The paper ends with a discussion of
the relationship between that account and incremental semantic
processing.

\section{Resources and Dependencies}
Most, if not all, contemporary linguistic theories are based on a view
of language in which linguistic entities produce and consume
resources.  For example, a transitive verb is standardly viewed as
something which requires two noun phrases, its subject and object, to
fill its ``thematic grid,'' i.e., fill its argument requirements in
order to produce a complete clause, which can in turn serve as an
argument of a higher verb.

Linguistic theories differ as to exactly what types of resources are
involved in particular sentences, what structural configurations are
required for a resource requirement to be filled, exactly what the
rules for resource accounting are, and what other mechanisms 
play a role in language.  For example, versions
of Categorial Grammar such as \namecite{Morrill94}'s Type-Logical
Grammar are based solely on (directionally sensitive) resource accounting.
On the other hand, modern versions of Transformational Grammar
specify a set of configurations
in which resource consumption can occur (e.g., feature cancellation
occurs in a Spec-Head configuration) and provides structural operations 
(Merge and Move) to move linguistic entities into such configurations.

The notion of ``incomplete dependency'' adopted in most processing
models derives from a fundamentally resource based view of language.
A dependency is a producer-consumer pair, and it is ``incomplete'' at
a point in a left to right analysis if exactly one member of the pair
has been encountered in the portion of the input string seen so far.
For example, at the point marked `\marker' in the simple transitive
sentence (\ref{e:trans}) the verb-object dependency is incomplete,
since the consumer resource (i.e., the verb) has been encountered but the
producer resource (the object noun phrase) has not.
\enumsentence{ \label{e:trans}
Kim kissed \marker\ Sandy.}
Recent work in logic has established a general framework in which
various kinds of resource sensitivity can be formalized and studied.
This section informally sketches some of the key ideas, but the reader
should see \namecite{Girard89} and \namecite{vanBenthem95} for a more
thorough introduction to this material.

Linguistic entities are classified into {\em types}, which encode
their combinatory capability.  For example, the noun phrase
{\em Kim} might be assigned the type $\NP$, and a verb
phrase {\em snores} might be assigned the type $\NP \uimp \S$,
indicating that it consumes a resource of type $\NP$ to produce
an object of type $\S$.  These types also indicate the {\em type}
of semantic interpretation associated with these phrases:
e.g., the verb phrase is a function from $\NP$ meanings to
$\S$ meanings.

Resource logics differ in terms of the kinds of {\em structural
sensitivity} they enforce.  For example, a grammar consisting solely
of a resource logic (e.g., Lambek Categorial Grammar) needs to be
sensitive to the positions of linguistic entities in the sentence.  In
such a grammar the location of the argument with respect to the
functor needs to be specified, so the undirected implication `$\uimp$'
is specialized into a leftward looking version `$\backslash$' and a
rightward looking version `$/$'.  For example, a transitive verb such
as {\em touches} might be of type $(\NP \backslash \S) / \NP$, i.e.,
an entity which consumes an $\NP$ to its right (the object noun
phrase) and then consumes an $\NP$ to its left (its subject) to
produce a saturated sentence.

Directional sensitivity is one dimension of structural sensitivity.
In general, the more linguistic detail a resource logic is called upon
to account for, the more refined its type system needs to be.  Because
the domain of locality of linguistic relationships may vary (e.g.,
head-complement dependencies are more local than \WH\ dependencies),
different types may need to be associated with different structural
sensitivities.  A general framework of multimodal substructural logics
has been developed for formalizing and investigating these
interactions \cite{Moortgat:Handbook}.

A proof in a resource logic specifies a ``plugging'', i.e., it
identifies which objects fill the requirements of which other objects
\cite{Girard95}.  These dependencies or pluggings determine the
semantic interpretation of the utterance; e.g., via the Curry-Howard
correspondence between (intuitionistic) proofs and $\lambda$-terms
\cite{Girard89,vanBenthem95}.  

The job of a grammar is to specify just which pluggings can occur in a
particular human language, and sophisticated resource logics have been
developed just for this purpose, as mentioned above.  However, for the
purposes of this paper it is not necessary to exactly identify the
constraints on grammatical pluggings: indeed, one of the strengths of
a dependency based approach is that it does not depend on the precise
details of the particular linguistic theory involved.  Rather, it
suffices that we can determine the pluggings which have actually
occured in the particular sentences being studied.  These pluggings
can usually be deduced from fairly general linguistic assumptions
(e.g., about the valence of verbs) and the interpretation of the
sentences themselves (which must reflect their pluggings, via the
Curry-Howard correspondence).

To keep things simple, all the examples in this paper use Lambek
Categorial Grammar.  This grammatical framework, while sufficient to
describe the examples presented here, is incapable of describing many
other important constructions that appear in natural languages.  More
sophisticated grammatical frameworks, such as the multi-modal systems
described \namecite{Morrill94} and \namecite{Moortgat:Handbook}, can
account for a much wider set of natural language examples.  However,
as the interested reader can confirm, the observations presented here
about the complexity of particular examples also hold in these more
complex systems (as they must, since these more sophisticated grammars
assume the same pluggings as presented here).

\subsection{Proof nets}
Proof nets are graphic depictions of proofs, i.e., of the
dependencies or plugging relationships between entities.  In the
application described here it is necessary to systematically
distinguish the inputs to the combinatory process (the lexical
items) from the single output produced from these inputs (the
completed sentence).  One way to do this is to use an intuitionistic
logic, and to use proof nets in which the edges are directed.

A proof net is a directed graph composed of proof net connectives.
The edges of the proof nets used here are directed (as the logic is
intuitionistic) and each edge is labelled with a type. A proof net
must satisfy structural conditions which depend on the kind of
structural sensitivity imposed by the logic.  Proof nets for Lambek
Categorial Grammars must satisfy (among other constraints) a planarity
condition \cite{Roorda91,Roorda92jlc,Lamarche96}.
\namecite{Moortgat:Handbook} describes in detail the structural
conditions that correspond to the more general class of multimodal
resource logics.  However, as explained above the work presented here
requires us only to identify the proof nets associated with particular
examples, and does not depend directly on the general structural
constraints themselves, so they are not presented here.

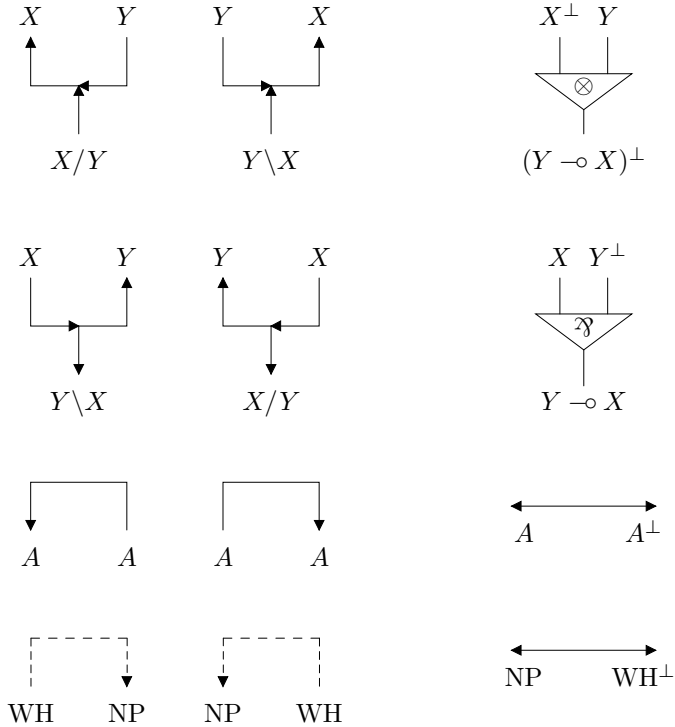
\begin{figure}
\begin{center}
\setlength{\unitlength}{0.00083333in}
\begingroup\makeatletter\ifx\SetFigFont\undefined%
\gdef\SetFigFont#1#2#3#4#5{%
  \reset@font\fontsize{#1}{#2pt}%
  \fontfamily{#3}\fontseries{#4}\fontshape{#5}%
  \selectfont}%
\fi\endgroup%
{\renewcommand{\dashlinestretch}{30}
\begin{picture}(4779,4548)(0,-10)
\dashline{60.000}(370,252)(370,552)(970,552)(970,252)
\blacken\path(940.000,312.000)(970.000,252.000)(1000.000,312.000)(940.000,312.000)
\put(370,27){\makebox(0,0)[b]{\smash{{{\SetFigFont{12}{14.4}{\rmdefault}{\mddefault}{\updefault}WH}}}}}
\put(970,27){\makebox(0,0)[b]{\smash{{{\SetFigFont{12}{14.4}{\rmdefault}{\mddefault}{\updefault}NP}}}}}
\dashline{60.000}(2170,252)(2170,552)(1570,552)(1570,252)
\blacken\path(1540.000,312.000)(1570.000,252.000)(1600.000,312.000)(1540.000,312.000)
\put(2170,27){\makebox(0,0)[b]{\smash{{{\SetFigFont{12}{14.4}{\rmdefault}{\mddefault}{\updefault}WH}}}}}
\put(1570,27){\makebox(0,0)[b]{\smash{{{\SetFigFont{12}{14.4}{\rmdefault}{\mddefault}{\updefault}NP}}}}}
\path(670,3702)(670,4002)
\blacken\path(700.000,3942.000)(670.000,4002.000)(640.000,3942.000)(700.000,3942.000)
\path(670,4002)(370,4002)(370,4302)
\blacken\path(400.000,4242.000)(370.000,4302.000)(340.000,4242.000)(400.000,4242.000)
\path(970,4302)(970,4002)(670,4002)
\blacken\path(730.000,4032.000)(670.000,4002.000)(730.000,3972.000)(730.000,4032.000)
\path(1570,4302)(1570,4002)(1870,4002)
\blacken\path(1810.000,3972.000)(1870.000,4002.000)(1810.000,4032.000)(1810.000,3972.000)
\path(1870,3702)(1870,4002)
\blacken\path(1900.000,3942.000)(1870.000,4002.000)(1840.000,3942.000)(1900.000,3942.000)
\path(1870,4002)(2170,4002)(2170,4302)
\blacken\path(2200.000,4242.000)(2170.000,4302.000)(2140.000,4242.000)(2200.000,4242.000)
\path(370,2802)(370,2502)(670,2502)
\blacken\path(610.000,2472.000)(670.000,2502.000)(610.000,2532.000)(610.000,2472.000)
\path(670,2502)(970,2502)(970,2802)
\blacken\path(1000.000,2742.000)(970.000,2802.000)(940.000,2742.000)(1000.000,2742.000)
\path(670,2502)(670,2202)
\blacken\path(640.000,2262.000)(670.000,2202.000)(700.000,2262.000)(640.000,2262.000)
\path(1870,2502)(1570,2502)(1570,2802)
\blacken\path(1600.000,2742.000)(1570.000,2802.000)(1540.000,2742.000)(1600.000,2742.000)
\path(2170,2802)(2170,2502)(1870,2502)
\blacken\path(1930.000,2532.000)(1870.000,2502.000)(1930.000,2472.000)(1930.000,2532.000)
\path(1870,2502)(1870,2202)
\blacken\path(1840.000,2262.000)(1870.000,2202.000)(1900.000,2262.000)(1840.000,2262.000)
\path(3820,2127)(3820,2352)
\path(3520,4077)(4120,4077)(3820,3852)(3520,4077)
\path(3820,3702)(3820,3852)
\path(3670,4077)(3670,4302)
\path(3970,4302)(3970,4077)
\path(3670,2577)(3670,2802)
\path(3970,2802)(3970,2577)
\path(3520,2577)(4120,2577)(3820,2352)(3520,2577)
\path(4270,477)(3370,477)
\blacken\path(4210.000,447.000)(4270.000,477.000)(4210.000,507.000)(4210.000,447.000)
\blacken\path(3430.000,507.000)(3370.000,477.000)(3430.000,447.000)(3430.000,507.000)
\path(970,1227)(970,1527)(370,1527)(370,1227)
\blacken\path(340.000,1287.000)(370.000,1227.000)(400.000,1287.000)(340.000,1287.000)
\path(1570,1227)(1570,1527)(2170,1527)(2170,1227)
\blacken\path(2140.000,1287.000)(2170.000,1227.000)(2200.000,1287.000)(2140.000,1287.000)
\path(4270,1377)(3370,1377)
\blacken\path(4210.000,1347.000)(4270.000,1377.000)(4210.000,1407.000)(4210.000,1347.000)
\blacken\path(3430.000,1407.000)(3370.000,1377.000)(3430.000,1347.000)(3430.000,1407.000)
\put(370,4377){\makebox(0,0)[b]{\smash{{{\SetFigFont{12}{14.4}{\rmdefault}{\mddefault}{\updefault}$X$}}}}}
\put(970,4377){\makebox(0,0)[b]{\smash{{{\SetFigFont{12}{14.4}{\rmdefault}{\mddefault}{\updefault}$Y$}}}}}
\put(670,3477){\makebox(0,0)[b]{\smash{{{\SetFigFont{12}{14.4}{\rmdefault}{\mddefault}{\updefault}$X/Y$}}}}}
\put(1570,4377){\makebox(0,0)[b]{\smash{{{\SetFigFont{12}{14.4}{\rmdefault}{\mddefault}{\updefault}$Y$}}}}}
\put(2170,4377){\makebox(0,0)[b]{\smash{{{\SetFigFont{12}{14.4}{\rmdefault}{\mddefault}{\updefault}$X$}}}}}
\put(1870,3477){\makebox(0,0)[b]{\smash{{{\SetFigFont{12}{14.4}{\rmdefault}{\mddefault}{\updefault}$Y\backslash X$}}}}}
\put(370,2877){\makebox(0,0)[b]{\smash{{{\SetFigFont{12}{14.4}{\rmdefault}{\mddefault}{\updefault}$X$}}}}}
\put(970,2877){\makebox(0,0)[b]{\smash{{{\SetFigFont{12}{14.4}{\rmdefault}{\mddefault}{\updefault}$Y$}}}}}
\put(670,1977){\makebox(0,0)[b]{\smash{{{\SetFigFont{12}{14.4}{\rmdefault}{\mddefault}{\updefault}$Y\backslash X$}}}}}
\put(1570,2877){\makebox(0,0)[b]{\smash{{{\SetFigFont{12}{14.4}{\rmdefault}{\mddefault}{\updefault}$Y$}}}}}
\put(2170,2877){\makebox(0,0)[b]{\smash{{{\SetFigFont{12}{14.4}{\rmdefault}{\mddefault}{\updefault}$X$}}}}}
\put(1870,1977){\makebox(0,0)[b]{\smash{{{\SetFigFont{12}{14.4}{\rmdefault}{\mddefault}{\updefault}$X/Y$}}}}}
\put(3820,2432){\makebox(0,0)[b]{\smash{{{\SetFigFont{12}{14.4}{\rmdefault}{\mddefault}{\updefault}$\lpar$}}}}}
\put(3820,3477){\makebox(0,0)[b]{\smash{{{\SetFigFont{12}{14.4}{\rmdefault}{\mddefault}{\updefault}$(Y\limp X)^\perp$}}}}}
\put(3820,1977){\makebox(0,0)[b]{\smash{{{\SetFigFont{12}{14.4}{\rmdefault}{\mddefault}{\updefault}$Y\limp X$}}}}}
\put(3820,3952){\makebox(0,0)[b]{\smash{{{\SetFigFont{12}{14.4}{\rmdefault}{\mddefault}{\updefault}$\otimes$}}}}}
\put(3970,4377){\makebox(0,0)[b]{\smash{{{\SetFigFont{12}{14.4}{\rmdefault}{\mddefault}{\updefault}$Y$}}}}}
\put(3670,4377){\makebox(0,0)[b]{\smash{{{\SetFigFont{12}{14.4}{\rmdefault}{\mddefault}{\updefault}$X^\perp$}}}}}
\put(3670,2877){\makebox(0,0)[b]{\smash{{{\SetFigFont{12}{14.4}{\rmdefault}{\mddefault}{\updefault}$X$}}}}}
\put(3970,2877){\makebox(0,0)[b]{\smash{{{\SetFigFont{12}{14.4}{\rmdefault}{\mddefault}{\updefault}$Y^\perp$}}}}}
\put(4195,252){\makebox(0,0)[b]{\smash{{{\SetFigFont{12}{14.4}{\rmdefault}{\mddefault}{\updefault}WH${}^\perp$}}}}}
\put(3445,252){\makebox(0,0)[b]{\smash{{{\SetFigFont{12}{14.4}{\rmdefault}{\mddefault}{\updefault}NP}}}}}
\put(970,1002){\makebox(0,0)[b]{\smash{{{\SetFigFont{12}{14.4}{\rmdefault}{\mddefault}{\updefault}$A$}}}}}
\put(370,1002){\makebox(0,0)[b]{\smash{{{\SetFigFont{12}{14.4}{\rmdefault}{\mddefault}{\updefault}$A$}}}}}
\put(1570,1002){\makebox(0,0)[b]{\smash{{{\SetFigFont{12}{14.4}{\rmdefault}{\mddefault}{\updefault}$A$}}}}}
\put(2170,1002){\makebox(0,0)[b]{\smash{{{\SetFigFont{12}{14.4}{\rmdefault}{\mddefault}{\updefault}$A$}}}}}
\put(4195,1152){\makebox(0,0)[b]{\smash{{{\SetFigFont{12}{14.4}{\rmdefault}{\mddefault}{\updefault}$A^\perp$}}}}}
\put(3445,1152){\makebox(0,0)[b]{\smash{{{\SetFigFont{12}{14.4}{\rmdefault}{\mddefault}{\updefault}$A$}}}}}
\end{picture}
}
\end{center}
\caption{ 
The first two columns list the proof net connectives used here, and 
the third column provides the corresponding linear logic proof 
net connective.  
The top two rows specify ``connector schema'' which
decompose complex types into their parts; outward-going arcs indicate
resources ``provided'' by this connector and inward-pointing arcs
indicate resources it ``consumes''.  The third row presents the axiom schema;
here $A$ ranges over atomic types.  The fourth row presents a non-logical
axiom permitting a \WH\ phrase to plug an \NP\ requirement.} \label{f:pn}
\end{figure}

The first two columns of Figure~\ref{f:pn} lists the proof net
connectives used in the examples here, and the third column lists the
corresponding linear logic proof net connective for readers familiar
with linear logic.  

We permit two kinds of axiom links.  In addition to the identity axiom
(the wire permitting an $A$ phrase to plug an $A$ requirement) we have
a non-logical axiom permitting a \WH\ phrase to plug an \NP\
requirement, and will indicate its use with a dashed line.  (This
distinction is one that any realistic grammar will have to make, as
\WH\ and \NP\ dependencies exhibit different structural constraints,
even though we do not make use of that fact here).

The proof net connectives decompose the types associated with
lexical entries into the resources that they produce and consume,
so each lexical entry can be regarded as a partial
proof net with incoming and outgoing edges labelled with atomic
categories.  These unpacked lexical entries are connected with other
words and phrases by axiom links.  

The axiom links identify the {\em dependencies} between the lexical
item and its surrounding material, so the unsatisfied dependencies at
any point in processing are precisely the axiom links that must be cut
in order to disconnect the proof net at the point of the
cut.\footnote{As an anonymous reviewer points out, in non-commutative
logics such as LCG a cyclic permutation property arises naturally from
one method of extending these logics to include negation, although it
is by no means necessary, as explained in \namecite{Lamarche96}. Such
a cyclic permutation of inputs can reduce the number of axiom links
that need to be cut in order to disconnect the proof net.  However,
these permutations of the input are not directly relevant to the
issues discussed here.  We are specifically interested in cuts that
divide the input into the words heard before a given point in time,
and those that come after that point in time, and in general a cut in
an arbitrary cyclic permutation of the inputs will have not have this
property.  

However, at a more general level this observation makes the important
point that non-incremental interpretation (i.e.  processing the input
in a different order to that in which it was perceived) may be more
economical than incremental interpretation, in the sense of involving
fewer incomplete dependencies during interpretation.  To my knowledge
this possibility has not been explored in the psycholinguistics
literature.}

\begin{figure}
\begin{center}
\setlength{\unitlength}{0.00083333in}
\begingroup\makeatletter\ifx\SetFigFont\undefined%
\gdef\SetFigFont#1#2#3#4#5{%
  \reset@font\fontsize{#1}{#2pt}%
  \fontfamily{#3}\fontseries{#4}\fontshape{#5}%
  \selectfont}%
\fi\endgroup%
{\renewcommand{\dashlinestretch}{30}
\begin{picture}(3246,2064)(0,-10)
\path(1785,537)(1785,687)
\blacken\path(1815.000,627.000)(1785.000,687.000)(1755.000,627.000)(1815.000,627.000)
\path(1485,912)(1485,687)(1785,687)
\blacken\path(1545.000,717.000)(1485.000,687.000)(1545.000,657.000)(1545.000,717.000)
\path(1785,687)(2085,687)(2085,912)
\blacken\path(2055.000,747.000)(2085.000,687.000)(2115.000,747.000)(2055.000,747.000)
\path(1185,1512)(1185,1287)(1485,1287)
\blacken\path(1425.000,1257.000)(1485.000,1287.000)(1425.000,1317.000)(1425.000,1257.000)
\path(1485,1137)(1485,1287)
\blacken\path(1515.000,1227.000)(1485.000,1287.000)(1455.000,1227.000)(1515.000,1227.000)
\path(1485,1287)(1785,1287)(1785,1512)
\blacken\path(1815.000,1452.000)(1785.000,1512.000)(1755.000,1452.000)(1815.000,1452.000)
\put(1785,387){\makebox(0,0)[b]{\smash{{{\SetFigFont{10}{12.0}{\familydefault}{\mddefault}{\updefault}(NP$\backslash$S)/NP}}}}}
\put(2085,987){\makebox(0,0)[b]{\smash{{{\SetFigFont{10}{12.0}{\familydefault}{\mddefault}{\updefault}NP}}}}}
\put(1785,1587){\makebox(0,0)[b]{\smash{{{\SetFigFont{10}{12.0}{\familydefault}{\mddefault}{\updefault}S}}}}}
\put(1485,987){\makebox(0,0)[b]{\smash{{{\SetFigFont{10}{12.0}{\familydefault}{\mddefault}{\updefault}NP$\backslash$S}}}}}
\put(1185,1587){\makebox(0,0)[b]{\smash{{{\SetFigFont{10}{12.0}{\familydefault}{\mddefault}{\updefault}NP}}}}}
\put(285,87){\makebox(0,0)[b]{\smash{{{\SetFigFont{10}{12.0}{\familydefault}{\mddefault}{\updefault}the nurse}}}}}
\put(285,387){\makebox(0,0)[b]{\smash{{{\SetFigFont{10}{12.0}{\familydefault}{\mddefault}{\updefault}NP}}}}}
\texture{88555555 55000000 555555 55000000 555555 55000000 555555 55000000 
	555555 55000000 555555 55000000 555555 55000000 555555 55000000 
	555555 55000000 555555 55000000 555555 55000000 555555 55000000 
	555555 55000000 555555 55000000 555555 55000000 555555 55000000 }
\shade\path(810,2037)(885,2037)(885,12)
	(810,12)(810,2037)
\path(810,2037)(885,2037)(885,12)
	(810,12)(810,2037)
\path(285,537)(285,1887)(1185,1887)(1185,1737)
\blacken\path(1155.000,1797.000)(1185.000,1737.000)(1215.000,1797.000)(1155.000,1797.000)
\shade\path(2385,2037)(2460,2037)(2460,12)
	(2385,12)(2385,2037)
\path(2385,2037)(2460,2037)(2460,12)
	(2385,12)(2385,2037)
\path(2985,537)(2985,1287)(2085,1287)(2085,1137)
\blacken\path(2055.000,1197.000)(2085.000,1137.000)(2115.000,1197.000)(2055.000,1197.000)
\put(1785,87){\makebox(0,0)[b]{\smash{{{\SetFigFont{10}{12.0}{\familydefault}{\mddefault}{\updefault}administered}}}}}
\put(2985,387){\makebox(0,0)[b]{\smash{{{\SetFigFont{10}{12.0}{\familydefault}{\mddefault}{\updefault}NP}}}}}
\put(2985,87){\makebox(0,0)[b]{\smash{{{\SetFigFont{10}{12.0}{\familydefault}{\mddefault}{\updefault}the drug}}}}}
\end{picture}
}
\end{center}
\caption{
A transitive clause, with vertical bars marking divisions between
the words.} \label{f:trans}
\end{figure}
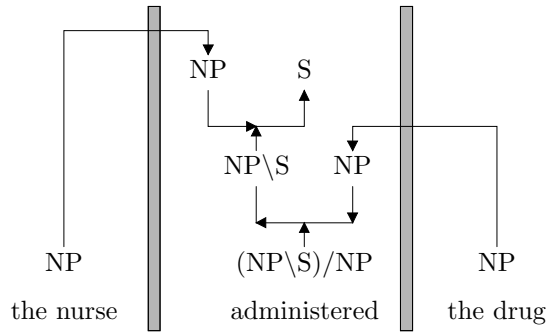

We end this section with some examples.  Figure~\ref{f:trans} shows
the proof net for the simple transitive sentence {\em Kim admires
Sandy}.  The vertical bars indicate cuts made between the words in
this sentence.  Each cut partitions the words or input resources into
two groups, corresponding to words received at some point of time in
processing the input.  Incremental processing in this framework
corresponds to the assumption that the input seen so far is integrated
as much as is possible, but the axiom links crossing a cut must be
disconnected at the point of time corresponding to that cut because
the resources they connect to have not yet been received.  In
example~\ref{f:trans} exactly one axiom link crosses each cut, so
there is exactly one unsatisfied \NP\ dependency at each of these
locations.

\begin{figure}
\begin{center}
\setlength{\unitlength}{0.00083333in}
\begingroup\makeatletter\ifx\SetFigFont\undefined%
\gdef\SetFigFont#1#2#3#4#5{%
  \reset@font\fontsize{#1}{#2pt}%
  \fontfamily{#3}\fontseries{#4}\fontshape{#5}%
  \selectfont}%
\fi\endgroup%
{\renewcommand{\dashlinestretch}{30}
\begin{picture}(3324,2462)(0,-10)
\path(12,1535)(12,1835)(612,1835)(612,1685)
\blacken\path(582.000,1745.000)(612.000,1685.000)(642.000,1745.000)(582.000,1745.000)
\path(3012,1535)(3012,1835)(2412,1835)(2412,1685)
\blacken\path(2382.000,1745.000)(2412.000,1685.000)(2442.000,1745.000)(2382.000,1745.000)
\dashline{60.000}(1812,1685)(1812,2135)(3312,2135)(3312,1535)
\blacken\path(3282.000,1595.000)(3312.000,1535.000)(3342.000,1595.000)(3282.000,1595.000)
\path(1512,471)(1512,621)
\blacken\path(1542.000,561.000)(1512.000,621.000)(1482.000,561.000)(1542.000,561.000)
\path(1512,621)(912,621)(912,846)
\blacken\path(942.000,786.000)(912.000,846.000)(882.000,786.000)(942.000,786.000)
\path(2112,846)(2112,621)(1512,621)
\blacken\path(1572.000,651.000)(1512.000,621.000)(1572.000,591.000)(1572.000,651.000)
\path(2112,1221)(2112,1071)
\blacken\path(2082.000,1131.000)(2112.000,1071.000)(2142.000,1131.000)(2082.000,1131.000)
\path(2412,1460)(2412,1235)(2112,1235)
\blacken\path(2172.000,1265.000)(2112.000,1235.000)(2172.000,1205.000)(2172.000,1265.000)
\path(912,1071)(912,1221)
\blacken\path(942.000,1161.000)(912.000,1221.000)(882.000,1161.000)(942.000,1161.000)
\path(612,1446)(612,1221)(912,1221)
\blacken\path(852.000,1191.000)(912.000,1221.000)(852.000,1251.000)(852.000,1191.000)
\path(912,1221)(1212,1221)(1212,1446)
\blacken\path(1242.000,1386.000)(1212.000,1446.000)(1182.000,1386.000)(1242.000,1386.000)
\path(2112,1235)(1812,1235)(1812,1460)
\blacken\path(1842.000,1400.000)(1812.000,1460.000)(1782.000,1400.000)(1842.000,1400.000)
\path(1212,1685)(1212,2435)
\blacken\path(1242.000,2375.000)(1212.000,2435.000)(1182.000,2375.000)(1242.000,2375.000)
\put(1512,21){\makebox(0,0)[b]{\smash{{{\SetFigFont{10}{12.0}{\familydefault}{\mddefault}{\updefault}who}}}}}
\put(1512,321){\makebox(0,0)[b]{\smash{{{\SetFigFont{10}{12.0}{\familydefault}{\mddefault}{\updefault}(NP$\backslash$NP)/(S/WH)}}}}}
\put(912,921){\makebox(0,0)[b]{\smash{{{\SetFigFont{10}{12.0}{\familydefault}{\mddefault}{\updefault}NP$\backslash$NP}}}}}
\put(2112,921){\makebox(0,0)[b]{\smash{{{\SetFigFont{10}{12.0}{\familydefault}{\mddefault}{\updefault}S/WH}}}}}
\put(612,1521){\makebox(0,0)[b]{\smash{{{\SetFigFont{10}{12.0}{\familydefault}{\mddefault}{\updefault}NP}}}}}
\put(1212,1521){\makebox(0,0)[b]{\smash{{{\SetFigFont{10}{12.0}{\familydefault}{\mddefault}{\updefault}NP}}}}}
\put(1812,1521){\makebox(0,0)[b]{\smash{{{\SetFigFont{10}{12.0}{\familydefault}{\mddefault}{\updefault}WH}}}}}
\put(2412,1521){\makebox(0,0)[b]{\smash{{{\SetFigFont{10}{12.0}{\familydefault}{\mddefault}{\updefault}S}}}}}
\end{picture}
}
\end{center}
\caption{
The partial proof net associated with an object relative pronoun 
{\em who}.  The partial proof net associated with a subject relative 
has the same structure except that the outgoing arc labelled \WH\ 
and the incoming arc labelled \S\ are swapped and type labels adjusted
accordingly.}
\label{f:who}
\end{figure}
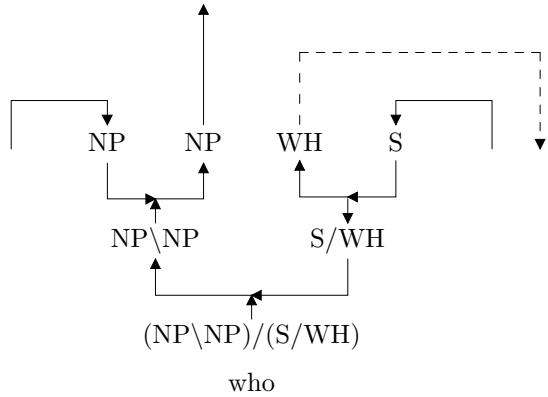

Figure~\ref{f:who} shows the partial proof net associated with a
relative pronoun.  The analysis of relative clauses used here is
simplified, and makes obviously incorrect linguistic predictions.  It
only permits peripheral \NP s to be ``extracted'', and does not
respect syntactic ``island'' constraints.  \NP s are treated as atomic
units in this paper, so relative clauses are analysed as \NP\
modifiers.  Thus a relative pronoun is analysed as something which
provides a \WH\ resource to its right and consumes the \S\ that it
appears in.  It also consumes an \NP\ to its left and produces a new
\NP, i.e., it functions as an \NP\ modifier.

These simplifications do not affect the analysis of the examples
presented in this paper, but clearly if other constructions were to be
analysed (such as medial extraction) a more sophisticated grammar
would be required.  \namecite{Morrill94} presents a detailed fragment
based on a multi-modal logic which permits clause medial extraction
and accounts for syntactic island phenomena.  His grammar assigns a
single type to relative pronouns, rather than the two types used
here. 
However, his grammar assigns proof structures to the examples
presented in this paper with the same general topological features 
as the ones presented here, and the interested reader can confirm
that the complexity differences presented below hold in his system
as well.

\section{Proof Nets, Dependencies and Processing} \label{s:pn}
With the formal tools of proof nets now available, it is relatively
simple to express many dependency-based theories of human sentence
processing complexity as geometric constraints on proof nets.

\begin{figure}
\setlength{\unitlength}{0.00083333in}
\begingroup\makeatletter\ifx\SetFigFont\undefined%
\gdef\SetFigFont#1#2#3#4#5{%
  \reset@font\fontsize{#1}{#2pt}%
  \fontfamily{#3}\fontseries{#4}\fontshape{#5}%
  \selectfont}%
\fi\endgroup%
{\renewcommand{\dashlinestretch}{30}
\begin{picture}(6248,2454)(0,-10)
\put(339,27){\makebox(0,0)[b]{\smash{{{\SetFigFont{10}{12.0}{\familydefault}{\mddefault}{\updefault}the patient}}}}}
\put(339,327){\makebox(0,0)[b]{\smash{{{\SetFigFont{10}{12.0}{\familydefault}{\mddefault}{\updefault}NP}}}}}
\put(2139,27){\makebox(0,0)[b]{\smash{{{\SetFigFont{10}{12.0}{\familydefault}{\mddefault}{\updefault}the drug}}}}}
\put(2139,327){\makebox(0,0)[b]{\smash{{{\SetFigFont{10}{12.0}{\familydefault}{\mddefault}{\updefault}NP}}}}}
\path(5139,477)(5139,627)
\blacken\path(5169.000,567.000)(5139.000,627.000)(5109.000,567.000)(5169.000,567.000)
\path(4839,852)(4839,627)(5139,627)
\blacken\path(4899.000,657.000)(4839.000,627.000)(4899.000,597.000)(4899.000,657.000)
\path(5139,627)(5439,627)(5439,852)
\blacken\path(5409.000,687.000)(5439.000,627.000)(5469.000,687.000)(5409.000,687.000)
\path(4539,1452)(4539,1227)(4839,1227)
\blacken\path(4779.000,1197.000)(4839.000,1227.000)(4779.000,1257.000)(4779.000,1197.000)
\path(4839,1077)(4839,1227)
\blacken\path(4869.000,1167.000)(4839.000,1227.000)(4809.000,1167.000)(4869.000,1167.000)
\path(4839,1227)(5139,1227)(5139,1452)
\blacken\path(5169.000,1392.000)(5139.000,1452.000)(5109.000,1392.000)(5169.000,1392.000)
\put(5139,327){\makebox(0,0)[b]{\smash{{{\SetFigFont{10}{12.0}{\familydefault}{\mddefault}{\updefault}(NP$\backslash$S)/NP}}}}}
\put(5439,927){\makebox(0,0)[b]{\smash{{{\SetFigFont{10}{12.0}{\familydefault}{\mddefault}{\updefault}NP}}}}}
\put(5139,1527){\makebox(0,0)[b]{\smash{{{\SetFigFont{10}{12.0}{\familydefault}{\mddefault}{\updefault}S}}}}}
\put(4839,927){\makebox(0,0)[b]{\smash{{{\SetFigFont{10}{12.0}{\familydefault}{\mddefault}{\updefault}NP$\backslash$S}}}}}
\put(4539,1527){\makebox(0,0)[b]{\smash{{{\SetFigFont{10}{12.0}{\familydefault}{\mddefault}{\updefault}NP}}}}}
\path(1539,477)(1539,627)
\blacken\path(1569.000,567.000)(1539.000,627.000)(1509.000,567.000)(1569.000,567.000)
\path(1239,852)(1239,627)(1539,627)
\blacken\path(1299.000,657.000)(1239.000,627.000)(1299.000,597.000)(1299.000,657.000)
\path(1539,627)(1839,627)(1839,852)
\blacken\path(1809.000,687.000)(1839.000,627.000)(1869.000,687.000)(1809.000,687.000)
\path(939,1452)(939,1227)(1239,1227)
\blacken\path(1179.000,1197.000)(1239.000,1227.000)(1179.000,1257.000)(1179.000,1197.000)
\path(1239,1077)(1239,1227)
\blacken\path(1269.000,1167.000)(1239.000,1227.000)(1209.000,1167.000)(1269.000,1167.000)
\path(1239,1227)(1539,1227)(1539,1452)
\blacken\path(1569.000,1392.000)(1539.000,1452.000)(1509.000,1392.000)(1569.000,1392.000)
\put(1539,327){\makebox(0,0)[b]{\smash{{{\SetFigFont{10}{12.0}{\familydefault}{\mddefault}{\updefault}(NP$\backslash$S)/NP}}}}}
\put(1839,927){\makebox(0,0)[b]{\smash{{{\SetFigFont{10}{12.0}{\familydefault}{\mddefault}{\updefault}NP}}}}}
\put(1539,1527){\makebox(0,0)[b]{\smash{{{\SetFigFont{10}{12.0}{\familydefault}{\mddefault}{\updefault}S}}}}}
\put(1239,927){\makebox(0,0)[b]{\smash{{{\SetFigFont{10}{12.0}{\familydefault}{\mddefault}{\updefault}NP$\backslash$S}}}}}
\put(939,1527){\makebox(0,0)[b]{\smash{{{\SetFigFont{10}{12.0}{\familydefault}{\mddefault}{\updefault}NP}}}}}
\path(339,477)(339,1827)(939,1827)(939,1677)
\blacken\path(909.000,1737.000)(939.000,1677.000)(969.000,1737.000)(909.000,1737.000)
\path(2139,477)(2139,1827)(2439,1827)(2439,1677)
\blacken\path(2409.000,1737.000)(2439.000,1677.000)(2469.000,1737.000)(2409.000,1737.000)
\path(3039,1677)(3039,2127)(1839,2127)(1839,1077)
\blacken\path(1809.000,1137.000)(1839.000,1077.000)(1869.000,1137.000)(1809.000,1137.000)
\path(5739,477)(5739,1227)(5439,1227)(5439,1077)
\blacken\path(5409.000,1137.000)(5439.000,1077.000)(5469.000,1137.000)(5409.000,1137.000)
\path(3339,477)(3339,627)
\blacken\path(3369.000,567.000)(3339.000,627.000)(3309.000,567.000)(3369.000,567.000)
\path(3339,627)(2739,627)(2739,852)
\blacken\path(2769.000,792.000)(2739.000,852.000)(2709.000,792.000)(2769.000,792.000)
\path(3939,852)(3939,627)(3339,627)
\blacken\path(3399.000,657.000)(3339.000,627.000)(3399.000,597.000)(3399.000,657.000)
\path(3939,1227)(3939,1077)
\blacken\path(3909.000,1137.000)(3939.000,1077.000)(3969.000,1137.000)(3909.000,1137.000)
\path(3939,1227)(4239,1227)(4239,1452)
\blacken\path(4269.000,1392.000)(4239.000,1452.000)(4209.000,1392.000)(4269.000,1392.000)
\path(3639,1452)(3639,1227)(3939,1227)
\blacken\path(3879.000,1197.000)(3939.000,1227.000)(3879.000,1257.000)(3879.000,1197.000)
\path(2739,1077)(2739,1227)
\blacken\path(2769.000,1167.000)(2739.000,1227.000)(2709.000,1167.000)(2769.000,1167.000)
\path(2439,1452)(2439,1227)(2739,1227)
\blacken\path(2679.000,1197.000)(2739.000,1227.000)(2679.000,1257.000)(2679.000,1197.000)
\path(2739,1227)(3039,1227)(3039,1452)
\blacken\path(3069.000,1392.000)(3039.000,1452.000)(3009.000,1392.000)(3069.000,1392.000)
\dashline{60.000}(4239,1677)(4239,1827)(4539,1827)(4539,1677)
\blacken\path(4509.000,1737.000)(4539.000,1677.000)(4569.000,1737.000)(4509.000,1737.000)
\path(5139,1677)(5139,2127)(3639,2127)(3639,1677)
\blacken\path(3609.000,1737.000)(3639.000,1677.000)(3669.000,1737.000)(3609.000,1737.000)
\path(1539,1677)(1539,2427)
\blacken\path(1569.000,2367.000)(1539.000,2427.000)(1509.000,2367.000)(1569.000,2367.000)
\put(5739,327){\makebox(0,0)[b]{\smash{{{\SetFigFont{10}{12.0}{\familydefault}{\mddefault}{\updefault}NP}}}}}
\put(3339,27){\makebox(0,0)[b]{\smash{{{\SetFigFont{10}{12.0}{\familydefault}{\mddefault}{\updefault}that}}}}}
\put(3339,327){\makebox(0,0)[b]{\smash{{{\SetFigFont{10}{12.0}{\familydefault}{\mddefault}{\updefault}(NP$\backslash$NP)/(WH$\backslash$S)}}}}}
\put(2739,927){\makebox(0,0)[b]{\smash{{{\SetFigFont{10}{12.0}{\familydefault}{\mddefault}{\updefault}NP$\backslash$NP}}}}}
\put(3939,927){\makebox(0,0)[b]{\smash{{{\SetFigFont{10}{12.0}{\familydefault}{\mddefault}{\updefault}WH$\backslash$S}}}}}
\put(2439,1527){\makebox(0,0)[b]{\smash{{{\SetFigFont{10}{12.0}{\familydefault}{\mddefault}{\updefault}NP}}}}}
\put(3039,1527){\makebox(0,0)[b]{\smash{{{\SetFigFont{10}{12.0}{\familydefault}{\mddefault}{\updefault}NP}}}}}
\put(3639,1527){\makebox(0,0)[b]{\smash{{{\SetFigFont{10}{12.0}{\familydefault}{\mddefault}{\updefault}S}}}}}
\put(4239,1527){\makebox(0,0)[b]{\smash{{{\SetFigFont{10}{12.0}{\familydefault}{\mddefault}{\updefault}WH}}}}}
\put(1314,27){\makebox(0,0)[b]{\smash{{{\SetFigFont{10}{12.0}{\familydefault}{\mddefault}{\updefault}was cured by}}}}}
\put(4839,27){\makebox(0,0)[b]{\smash{{{\SetFigFont{10}{12.0}{\familydefault}{\mddefault}{\updefault}was administered by}}}}}
\put(5964,27){\makebox(0,0)[b]{\smash{{{\SetFigFont{10}{12.0}{\familydefault}{\mddefault}{\updefault}the nurse}}}}}
\end{picture}
}
\caption{
A proof net for a relative clause modifying the object.}
\label{f:right-embedding}
\end{figure}
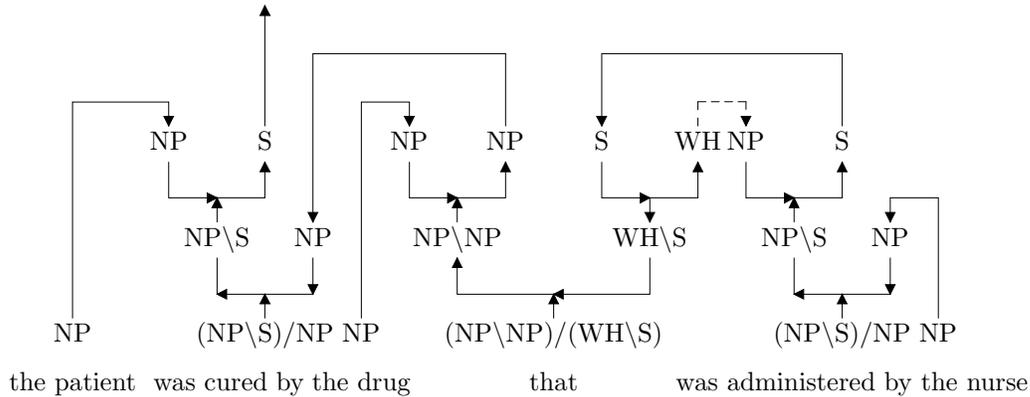

First, consider the iterated object relative clauses shown in (\ref{e:objrel}),
which are usually described as right-branching constructions.  While 
increased length does reduce acceptability, there does not seem to be
any significant processing overload associated with such examples.

\eenumsentence{      \label{e:objrel} 
\item The patient was cured by the drug [ that was administered by the nurse ].
     \label{e:objrela}
\item The patient was cured by the drug [ that was administered by the nurse
      [ who was supervised by the doctor ] ].
\item The patient was cured by the drug [ that was administered by the nurse
      [ who was supervised by the doctor [ who was admired by the student ] ].
}

Figure~\ref{f:right-embedding} depicts the proof net for
(\ref{e:objrela}).  For simplicity in this proof net the passivised
forms {\em was cured by} and {\em was administered by} are treated as
transitive verbs, rather than being analysed into their component
lexical items.  At most 2~axiom links cross any cut in the proof
nets for these examples (either two \NP\ links or an \NP\ and a \WH\
link, depending on where the cut lies), which is consistent with
processing complexity being independent of the number of such
constructions involved.

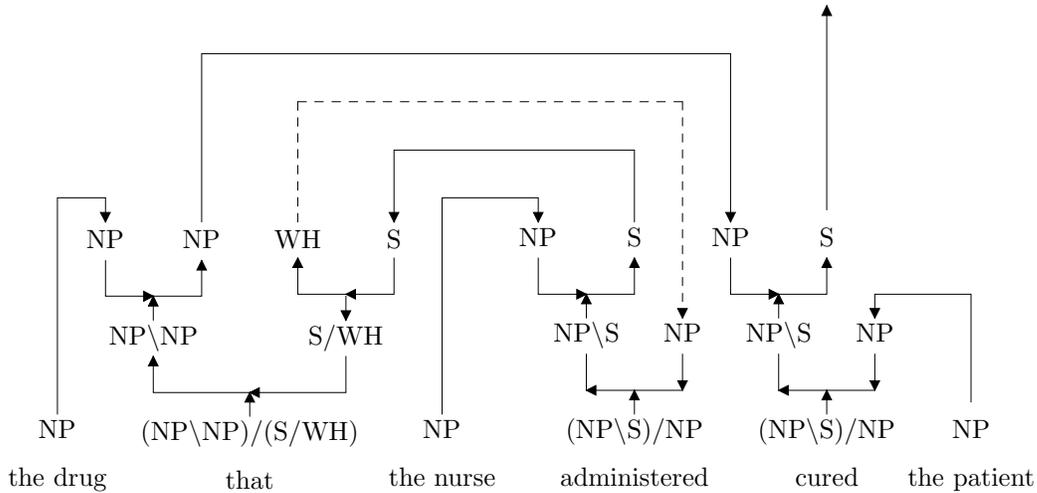
\begin{figure}
\setlength{\unitlength}{0.00083333in}
\begingroup\makeatletter\ifx\SetFigFont\undefined%
\gdef\SetFigFont#1#2#3#4#5{%
  \reset@font\fontsize{#1}{#2pt}%
  \fontfamily{#3}\fontseries{#4}\fontshape{#5}%
  \selectfont}%
\fi\endgroup%
{\renewcommand{\dashlinestretch}{30}
\begin{picture}(6300,3062)(0,-10)
\put(262,35){\makebox(0,0)[b]{\smash{{{\SetFigFont{10}{12.0}{\familydefault}{\mddefault}{\updefault}the drug}}}}}
\put(262,335){\makebox(0,0)[b]{\smash{{{\SetFigFont{10}{12.0}{\familydefault}{\mddefault}{\updefault}NP}}}}}
\put(2662,35){\makebox(0,0)[b]{\smash{{{\SetFigFont{10}{12.0}{\familydefault}{\mddefault}{\updefault}the nurse}}}}}
\put(2662,335){\makebox(0,0)[b]{\smash{{{\SetFigFont{10}{12.0}{\familydefault}{\mddefault}{\updefault}NP}}}}}
\path(3862,485)(3862,635)
\blacken\path(3892.000,575.000)(3862.000,635.000)(3832.000,575.000)(3892.000,575.000)
\path(3562,860)(3562,635)(3862,635)
\blacken\path(3622.000,665.000)(3562.000,635.000)(3622.000,605.000)(3622.000,665.000)
\path(3862,635)(4162,635)(4162,860)
\blacken\path(4132.000,695.000)(4162.000,635.000)(4192.000,695.000)(4132.000,695.000)
\path(3262,1460)(3262,1235)(3562,1235)
\blacken\path(3502.000,1205.000)(3562.000,1235.000)(3502.000,1265.000)(3502.000,1205.000)
\path(3562,1085)(3562,1235)
\blacken\path(3592.000,1175.000)(3562.000,1235.000)(3532.000,1175.000)(3592.000,1175.000)
\path(3562,1235)(3862,1235)(3862,1460)
\blacken\path(3892.000,1400.000)(3862.000,1460.000)(3832.000,1400.000)(3892.000,1400.000)
\put(3862,335){\makebox(0,0)[b]{\smash{{{\SetFigFont{10}{12.0}{\familydefault}{\mddefault}{\updefault}(NP$\backslash$S)/NP}}}}}
\put(4162,935){\makebox(0,0)[b]{\smash{{{\SetFigFont{10}{12.0}{\familydefault}{\mddefault}{\updefault}NP}}}}}
\put(3862,1535){\makebox(0,0)[b]{\smash{{{\SetFigFont{10}{12.0}{\familydefault}{\mddefault}{\updefault}S}}}}}
\put(3562,935){\makebox(0,0)[b]{\smash{{{\SetFigFont{10}{12.0}{\familydefault}{\mddefault}{\updefault}NP$\backslash$S}}}}}
\put(3262,1535){\makebox(0,0)[b]{\smash{{{\SetFigFont{10}{12.0}{\familydefault}{\mddefault}{\updefault}NP}}}}}
\path(5062,485)(5062,635)
\blacken\path(5092.000,575.000)(5062.000,635.000)(5032.000,575.000)(5092.000,575.000)
\path(4762,860)(4762,635)(5062,635)
\blacken\path(4822.000,665.000)(4762.000,635.000)(4822.000,605.000)(4822.000,665.000)
\path(5062,635)(5362,635)(5362,860)
\blacken\path(5332.000,695.000)(5362.000,635.000)(5392.000,695.000)(5332.000,695.000)
\path(4462,1460)(4462,1235)(4762,1235)
\blacken\path(4702.000,1205.000)(4762.000,1235.000)(4702.000,1265.000)(4702.000,1205.000)
\path(4762,1085)(4762,1235)
\blacken\path(4792.000,1175.000)(4762.000,1235.000)(4732.000,1175.000)(4792.000,1175.000)
\path(4762,1235)(5062,1235)(5062,1460)
\blacken\path(5092.000,1400.000)(5062.000,1460.000)(5032.000,1400.000)(5092.000,1400.000)
\put(5062,335){\makebox(0,0)[b]{\smash{{{\SetFigFont{10}{12.0}{\familydefault}{\mddefault}{\updefault}(NP$\backslash$S)/NP}}}}}
\put(5362,935){\makebox(0,0)[b]{\smash{{{\SetFigFont{10}{12.0}{\familydefault}{\mddefault}{\updefault}NP}}}}}
\put(5062,1535){\makebox(0,0)[b]{\smash{{{\SetFigFont{10}{12.0}{\familydefault}{\mddefault}{\updefault}S}}}}}
\put(4762,935){\makebox(0,0)[b]{\smash{{{\SetFigFont{10}{12.0}{\familydefault}{\mddefault}{\updefault}NP$\backslash$S}}}}}
\put(4462,1535){\makebox(0,0)[b]{\smash{{{\SetFigFont{10}{12.0}{\familydefault}{\mddefault}{\updefault}NP}}}}}
\path(1162,1685)(1162,2735)(4462,2735)(4462,1685)
\blacken\path(4432.000,1745.000)(4462.000,1685.000)(4492.000,1745.000)(4432.000,1745.000)
\path(1462,471)(1462,621)
\blacken\path(1492.000,561.000)(1462.000,621.000)(1432.000,561.000)(1492.000,561.000)
\path(1462,621)(862,621)(862,846)
\blacken\path(892.000,786.000)(862.000,846.000)(832.000,786.000)(892.000,786.000)
\path(2062,846)(2062,621)(1462,621)
\blacken\path(1522.000,651.000)(1462.000,621.000)(1522.000,591.000)(1522.000,651.000)
\path(2062,1221)(2062,1071)
\blacken\path(2032.000,1131.000)(2062.000,1071.000)(2092.000,1131.000)(2032.000,1131.000)
\path(2362,1460)(2362,1235)(2062,1235)
\blacken\path(2122.000,1265.000)(2062.000,1235.000)(2122.000,1205.000)(2122.000,1265.000)
\path(862,1071)(862,1221)
\blacken\path(892.000,1161.000)(862.000,1221.000)(832.000,1161.000)(892.000,1161.000)
\path(562,1446)(562,1221)(862,1221)
\blacken\path(802.000,1191.000)(862.000,1221.000)(802.000,1251.000)(802.000,1191.000)
\path(862,1221)(1162,1221)(1162,1446)
\blacken\path(1192.000,1386.000)(1162.000,1446.000)(1132.000,1386.000)(1192.000,1386.000)
\path(2062,1235)(1762,1235)(1762,1460)
\blacken\path(1792.000,1400.000)(1762.000,1460.000)(1732.000,1400.000)(1792.000,1400.000)
\path(262,485)(262,1835)(562,1835)(562,1685)
\blacken\path(532.000,1745.000)(562.000,1685.000)(592.000,1745.000)(532.000,1745.000)
\path(3862,1685)(3862,2135)(2362,2135)(2362,1685)
\blacken\path(2332.000,1745.000)(2362.000,1685.000)(2392.000,1745.000)(2332.000,1745.000)
\dashline{60.000}(1762,1685)(1762,2435)(4162,2435)(4162,1085)
\blacken\path(4132.000,1145.000)(4162.000,1085.000)(4192.000,1145.000)(4132.000,1145.000)
\path(5962,560)(5962,1235)(5362,1235)(5362,1085)
\blacken\path(5332.000,1145.000)(5362.000,1085.000)(5392.000,1145.000)(5332.000,1145.000)
\path(5062,1760)(5062,3035)
\blacken\path(5092.000,2975.000)(5062.000,3035.000)(5032.000,2975.000)(5092.000,2975.000)
\path(2662,485)(2662,1835)(3262,1835)(3262,1685)
\blacken\path(3232.000,1745.000)(3262.000,1685.000)(3292.000,1745.000)(3232.000,1745.000)
\put(1462,21){\makebox(0,0)[b]{\smash{{{\SetFigFont{10}{12.0}{\familydefault}{\mddefault}{\updefault}that}}}}}
\put(1462,321){\makebox(0,0)[b]{\smash{{{\SetFigFont{10}{12.0}{\familydefault}{\mddefault}{\updefault}(NP$\backslash$NP)/(S/WH)}}}}}
\put(862,921){\makebox(0,0)[b]{\smash{{{\SetFigFont{10}{12.0}{\familydefault}{\mddefault}{\updefault}NP$\backslash$NP}}}}}
\put(2062,921){\makebox(0,0)[b]{\smash{{{\SetFigFont{10}{12.0}{\familydefault}{\mddefault}{\updefault}S/WH}}}}}
\put(562,1521){\makebox(0,0)[b]{\smash{{{\SetFigFont{10}{12.0}{\familydefault}{\mddefault}{\updefault}NP}}}}}
\put(1162,1521){\makebox(0,0)[b]{\smash{{{\SetFigFont{10}{12.0}{\familydefault}{\mddefault}{\updefault}NP}}}}}
\put(1762,1521){\makebox(0,0)[b]{\smash{{{\SetFigFont{10}{12.0}{\familydefault}{\mddefault}{\updefault}WH}}}}}
\put(2362,1521){\makebox(0,0)[b]{\smash{{{\SetFigFont{10}{12.0}{\familydefault}{\mddefault}{\updefault}S}}}}}
\put(5962,35){\makebox(0,0)[b]{\smash{{{\SetFigFont{10}{12.0}{\familydefault}{\mddefault}{\updefault}the patient}}}}}
\put(5962,335){\makebox(0,0)[b]{\smash{{{\SetFigFont{10}{12.0}{\familydefault}{\mddefault}{\updefault}NP}}}}}
\put(3862,35){\makebox(0,0)[b]{\smash{{{\SetFigFont{10}{12.0}{\familydefault}{\mddefault}{\updefault}administered}}}}}
\put(5062,35){\makebox(0,0)[b]{\smash{{{\SetFigFont{10}{12.0}{\familydefault}{\mddefault}{\updefault}cured}}}}}
\end{picture}
}
\caption{A proof net for a relative clause modifying the subject.}
\label{f:rc-subj}
\end{figure}

Now consider the iterated subject relative clauses shown in (\ref{e:subjrel}),
which are classic examples of center-embedded constructions.  In contrast
to the object relative clauses in (\ref{e:objrel}), two or more levels of
embedding renders the example virtually incomprehensible.

\eenumsentence{ \label{e:subjrel}
\item The drug [ that the nurse administered ] cured the patient.
      \label{e:subjrela}
\item \bad The drug [ that the nurse [ that the doctor supervised ]
      administered ] cured the patient.
\item \bad The drug [ that the nurse [ that the doctor [ that the
      students admired ] supervised ] administered ] cured the patient.
}

Figure~\ref{f:rc-subj} depicts the proof net for (\ref{e:subjrela}).
The maximal cut in the constructions (\ref{e:subjrel}) occurs just
after the most embedded subject.  At depth of embedding $n$ the
maximal cut crosses $3n+1$ axiom links.  The assumption that the human
language processor can only keep track of a small number of such
incomplete dependencies accounts for the increasing ill-formedness as
the number of such constructions increases.

\begin{figure}
\begin{center}
\setlength{\unitlength}{0.00083333in}
\begingroup\makeatletter\ifx\SetFigFont\undefined%
\gdef\SetFigFont#1#2#3#4#5{%
  \reset@font\fontsize{#1}{#2pt}%
  \fontfamily{#3}\fontseries{#4}\fontshape{#5}%
  \selectfont}%
\fi\endgroup%
{\renewcommand{\dashlinestretch}{30}
\begin{picture}(5642,2754)(0,-10)
\path(408,477)(408,627)
\blacken\path(438.000,567.000)(408.000,627.000)(378.000,567.000)(438.000,567.000)
\path(708,852)(708,627)(408,627)
\blacken\path(468.000,657.000)(408.000,627.000)(468.000,597.000)(468.000,657.000)
\path(408,627)(108,627)(108,852)
\blacken\path(138.000,792.000)(108.000,852.000)(78.000,792.000)(138.000,792.000)
\path(1308,477)(1308,627)
\blacken\path(1338.000,567.000)(1308.000,627.000)(1278.000,567.000)(1338.000,567.000)
\path(1608,852)(1608,627)(1308,627)
\blacken\path(1368.000,657.000)(1308.000,627.000)(1368.000,597.000)(1368.000,657.000)
\path(1308,627)(1008,627)(1008,852)
\blacken\path(1038.000,792.000)(1008.000,852.000)(978.000,792.000)(1038.000,792.000)
\path(2808,1077)(2808,1227)
\blacken\path(2838.000,1167.000)(2808.000,1227.000)(2778.000,1167.000)(2838.000,1167.000)
\path(2808,1227)(3108,1227)(3108,1452)
\blacken\path(3138.000,1392.000)(3108.000,1452.000)(3078.000,1392.000)(3138.000,1392.000)
\path(2508,1452)(2508,1227)(2808,1227)
\blacken\path(2748.000,1197.000)(2808.000,1227.000)(2748.000,1257.000)(2748.000,1197.000)
\put(2808,927){\makebox(0,0)[b]{\smash{{{\SetFigFont{10}{12.0}{\rmdefault}{\mddefault}{\updefault}NP$\backslash$S}}}}}
\path(4308,1077)(4308,1227)
\blacken\path(4338.000,1167.000)(4308.000,1227.000)(4278.000,1167.000)(4338.000,1167.000)
\path(4308,1227)(4608,1227)(4608,1452)
\blacken\path(4638.000,1392.000)(4608.000,1452.000)(4578.000,1392.000)(4638.000,1392.000)
\path(4008,1452)(4008,1227)(4308,1227)
\blacken\path(4248.000,1197.000)(4308.000,1227.000)(4248.000,1257.000)(4248.000,1197.000)
\put(4308,927){\makebox(0,0)[b]{\smash{{{\SetFigFont{10}{12.0}{\rmdefault}{\mddefault}{\updefault}NP$\backslash$S}}}}}
\path(4608,477)(4608,627)
\blacken\path(4638.000,567.000)(4608.000,627.000)(4578.000,567.000)(4638.000,567.000)
\path(4908,852)(4908,627)(4608,627)
\blacken\path(4668.000,657.000)(4608.000,627.000)(4668.000,597.000)(4668.000,657.000)
\path(4608,627)(4308,627)(4308,852)
\blacken\path(4338.000,792.000)(4308.000,852.000)(4278.000,792.000)(4338.000,792.000)
\path(1908,477)(1908,1827)(2508,1827)(2508,1677)
\blacken\path(2478.000,1737.000)(2508.000,1677.000)(2538.000,1737.000)(2478.000,1737.000)
\path(1008,1077)(1008,1227)(708,1227)(708,1077)
\blacken\path(678.000,1137.000)(708.000,1077.000)(738.000,1137.000)(678.000,1137.000)
\path(3108,1677)(3108,2127)(1608,2127)(1608,1077)
\blacken\path(1578.000,1137.000)(1608.000,1077.000)(1638.000,1137.000)(1578.000,1137.000)
\path(3708,477)(3708,1227)(3408,1227)(3408,1077)
\blacken\path(3378.000,1137.000)(3408.000,1077.000)(3438.000,1137.000)(3378.000,1137.000)
\path(108,1077)(108,2427)(4008,2427)(4008,1677)
\blacken\path(3978.000,1737.000)(4008.000,1677.000)(4038.000,1737.000)(3978.000,1737.000)
\path(5208,477)(5208,1227)(4908,1227)(4908,1077)
\blacken\path(4878.000,1137.000)(4908.000,1077.000)(4938.000,1137.000)(4878.000,1137.000)
\path(4608,1677)(4608,2727)
\blacken\path(4638.000,2667.000)(4608.000,2727.000)(4578.000,2667.000)(4638.000,2667.000)
\path(3108,477)(3108,627)
\blacken\path(3138.000,567.000)(3108.000,627.000)(3078.000,567.000)(3138.000,567.000)
\path(3408,852)(3408,627)(3108,627)
\blacken\path(3168.000,657.000)(3108.000,627.000)(3168.000,597.000)(3168.000,657.000)
\path(3108,627)(2808,627)(2808,852)
\blacken\path(2838.000,792.000)(2808.000,852.000)(2778.000,792.000)(2838.000,792.000)
\put(408,27){\makebox(0,0)[b]{\smash{{{\SetFigFont{10}{12.0}{\rmdefault}{\mddefault}{\updefault}the chance}}}}}
\put(408,327){\makebox(0,0)[b]{\smash{{{\SetFigFont{10}{12.0}{\rmdefault}{\mddefault}{\updefault}NP/S'}}}}}
\put(108,927){\makebox(0,0)[b]{\smash{{{\SetFigFont{10}{12.0}{\rmdefault}{\mddefault}{\updefault}NP}}}}}
\put(708,927){\makebox(0,0)[b]{\smash{{{\SetFigFont{10}{12.0}{\rmdefault}{\mddefault}{\updefault}S'}}}}}
\put(1008,927){\makebox(0,0)[b]{\smash{{{\SetFigFont{10}{12.0}{\rmdefault}{\mddefault}{\updefault}S'}}}}}
\put(1608,927){\makebox(0,0)[b]{\smash{{{\SetFigFont{10}{12.0}{\rmdefault}{\mddefault}{\updefault}S}}}}}
\put(1308,327){\makebox(0,0)[b]{\smash{{{\SetFigFont{10}{12.0}{\rmdefault}{\mddefault}{\updefault}S'/S}}}}}
\put(1308,27){\makebox(0,0)[b]{\smash{{{\SetFigFont{10}{12.0}{\rmdefault}{\mddefault}{\updefault}that}}}}}
\put(1908,27){\makebox(0,0)[b]{\smash{{{\SetFigFont{10}{12.0}{\rmdefault}{\mddefault}{\updefault}the doctor}}}}}
\put(1908,327){\makebox(0,0)[b]{\smash{{{\SetFigFont{10}{12.0}{\rmdefault}{\mddefault}{\updefault}NP}}}}}
\put(2508,1527){\makebox(0,0)[b]{\smash{{{\SetFigFont{10}{12.0}{\rmdefault}{\mddefault}{\updefault}NP}}}}}
\put(3108,1527){\makebox(0,0)[b]{\smash{{{\SetFigFont{10}{12.0}{\rmdefault}{\mddefault}{\updefault}S}}}}}
\put(3108,327){\makebox(0,0)[b]{\smash{{{\SetFigFont{10}{12.0}{\rmdefault}{\mddefault}{\updefault}(NP$\backslash$S)/NP}}}}}
\put(3408,927){\makebox(0,0)[b]{\smash{{{\SetFigFont{10}{12.0}{\rmdefault}{\mddefault}{\updefault}NP}}}}}
\put(3108,27){\makebox(0,0)[b]{\smash{{{\SetFigFont{10}{12.0}{\rmdefault}{\mddefault}{\updefault}lost}}}}}
\put(3708,327){\makebox(0,0)[b]{\smash{{{\SetFigFont{10}{12.0}{\rmdefault}{\mddefault}{\updefault}NP}}}}}
\put(3708,27){\makebox(0,0)[b]{\smash{{{\SetFigFont{10}{12.0}{\rmdefault}{\mddefault}{\updefault}the reports}}}}}
\put(4008,1527){\makebox(0,0)[b]{\smash{{{\SetFigFont{10}{12.0}{\rmdefault}{\mddefault}{\updefault}NP}}}}}
\put(4608,1527){\makebox(0,0)[b]{\smash{{{\SetFigFont{10}{12.0}{\rmdefault}{\mddefault}{\updefault}S}}}}}
\put(4308,927){\makebox(0,0)[b]{\smash{{{\SetFigFont{10}{12.0}{\rmdefault}{\mddefault}{\updefault}NP$\backslash$S}}}}}
\put(4608,327){\makebox(0,0)[b]{\smash{{{\SetFigFont{10}{12.0}{\rmdefault}{\mddefault}{\updefault}(NP$\backslash$S)/NP}}}}}
\put(4908,927){\makebox(0,0)[b]{\smash{{{\SetFigFont{10}{12.0}{\rmdefault}{\mddefault}{\updefault}NP}}}}}
\put(4608,27){\makebox(0,0)[b]{\smash{{{\SetFigFont{10}{12.0}{\rmdefault}{\mddefault}{\updefault}bothered}}}}}
\put(5208,327){\makebox(0,0)[b]{\smash{{{\SetFigFont{10}{12.0}{\rmdefault}{\mddefault}{\updefault}NP}}}}}
\put(5358,27){\makebox(0,0)[b]{\smash{{{\SetFigFont{10}{12.0}{\rmdefault}{\mddefault}{\updefault}the nurse}}}}}
\end{picture}
}
\end{center}
\caption{A proof net for a complement subject example}
\label{f:sentsubj}
\end{figure}
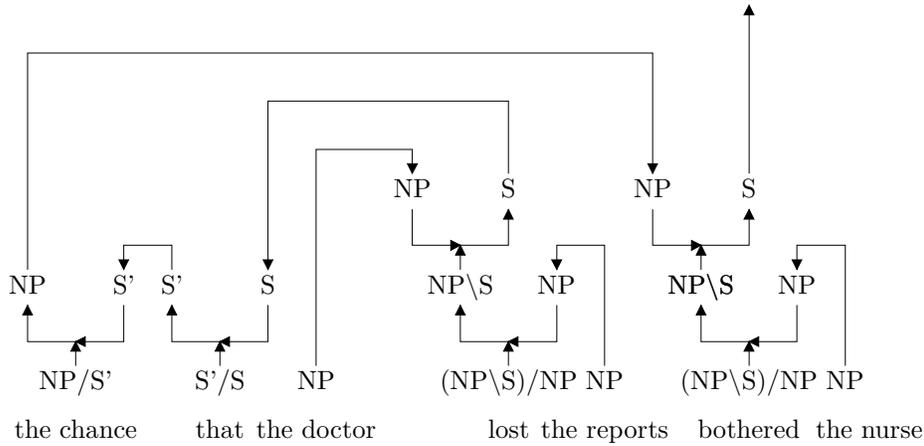

It is sometimes claimed that subject complement clauses such as
(\ref{e:sentsubj}) are easier to comprehend than corresponding subject relatives
\cite{Stabler94}.  Figure~\ref{f:sentsubj} depicts the proof net for
(\ref{e:sentsubj}).  The maximal cut for this example crosses one less
axiom link than the corresponding relative clause example
(\ref{e:subjrela}), which is consistent with this
putative difference.

\enumsentence{ \label{e:sentsubj}
The chance [ that the doctor lost the reports ] bothered the nurse.
}

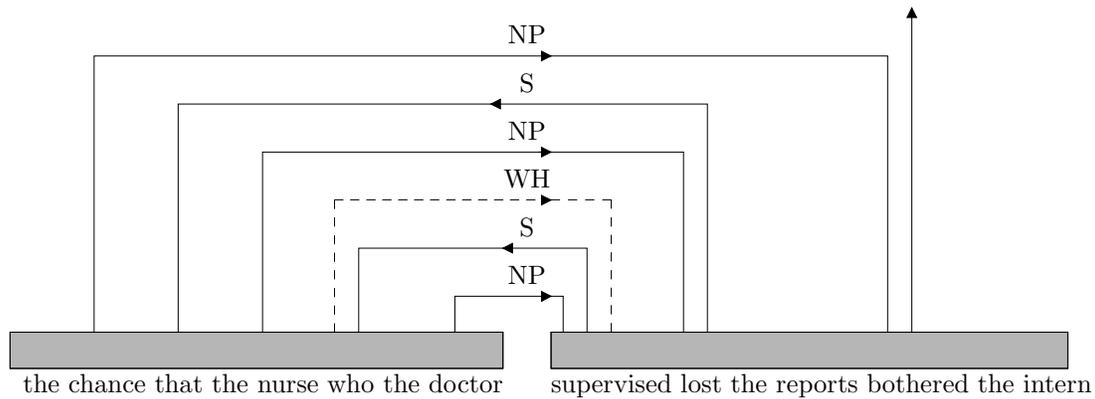
\begin{figure}
\setlength{\unitlength}{0.00083333in}
\begingroup\makeatletter\ifx\SetFigFont\undefined%
\gdef\SetFigFont#1#2#3#4#5{%
  \reset@font\fontsize{#1}{#2pt}%
  \fontfamily{#3}\fontseries{#4}\fontshape{#5}%
  \selectfont}%
\fi\endgroup%
{\renewcommand{\dashlinestretch}{30}
\begin{picture}(7007,2463)(0,-10)
\path(2823,411)(2823,636)(3423,636)
\blacken\path(3363.000,606.000)(3423.000,636.000)(3363.000,666.000)(3363.000,606.000)
\path(3423,636)(3498,636)(3498,411)
\path(3648,411)(3648,936)(3123,936)
\blacken\path(3183.000,966.000)(3123.000,936.000)(3183.000,906.000)(3183.000,966.000)
\path(3123,936)(2223,936)(2223,411)
\dashline{60.000}(3423,1236)(3798,1236)(3798,411)
\path(573,411)(573,2136)(3423,2136)
\blacken\path(3363.000,2106.000)(3423.000,2136.000)(3363.000,2166.000)(3363.000,2106.000)
\path(3423,2136)(5523,2136)(5523,411)
\texture{88555555 55000000 555555 55000000 555555 55000000 555555 55000000 
	555555 55000000 555555 55000000 555555 55000000 555555 55000000 
	555555 55000000 555555 55000000 555555 55000000 555555 55000000 
	555555 55000000 555555 55000000 555555 55000000 555555 55000000 }
\shade\path(3123,186)(48,186)(48,411)
	(3123,411)(3123,186)
\path(3123,186)(48,186)(48,411)
	(3123,411)(3123,186)
\path(5673,411)(5673,2436)
\blacken\path(5703.000,2376.000)(5673.000,2436.000)(5643.000,2376.000)(5703.000,2376.000)
\shade\path(3423,186)(6648,186)(6648,411)
	(3423,411)(3423,186)
\path(3423,186)(6648,186)(6648,411)
	(3423,411)(3423,186)
\dashline{60.000}(2073,411)(2073,1236)(3423,1236)
\blacken\path(3363.000,1206.000)(3423.000,1236.000)(3363.000,1266.000)(3363.000,1206.000)
\path(3048,1836)(1098,1836)(1098,411)
\path(1623,411)(1623,1536)(3423,1536)
\blacken\path(3363.000,1506.000)(3423.000,1536.000)(3363.000,1566.000)(3363.000,1506.000)
\path(4398,411)(4398,1836)(3048,1836)
\blacken\path(3108.000,1866.000)(3048.000,1836.000)(3108.000,1806.000)(3108.000,1866.000)
\path(3348,1536)(4248,1536)(4248,411)
\put(3123,36){\makebox(0,0)[rb]{\smash{{{\SetFigFont{12}{14.4}{\familydefault}{\mddefault}{\updefault}the chance that the nurse who the doctor}}}}}
\put(3423,36){\makebox(0,0)[lb]{\smash{{{\SetFigFont{12}{14.4}{\familydefault}{\mddefault}{\updefault}supervised lost the reports bothered the intern}}}}}
\put(3273,711){\makebox(0,0)[b]{\smash{{{\SetFigFont{12}{14.4}{\familydefault}{\mddefault}{\updefault}NP}}}}}
\put(3273,1011){\makebox(0,0)[b]{\smash{{{\SetFigFont{12}{14.4}{\familydefault}{\mddefault}{\updefault}S}}}}}
\put(3273,2211){\makebox(0,0)[b]{\smash{{{\SetFigFont{12}{14.4}{\familydefault}{\mddefault}{\updefault}NP}}}}}
\put(3273,1311){\makebox(0,0)[b]{\smash{{{\SetFigFont{12}{14.4}{\familydefault}{\mddefault}{\updefault}WH}}}}}
\put(3273,1911){\makebox(0,0)[b]{\smash{{{\SetFigFont{12}{14.4}{\familydefault}{\mddefault}{\updefault}S}}}}}
\put(3273,1611){\makebox(0,0)[b]{\smash{{{\SetFigFont{12}{14.4}{\familydefault}{\mddefault}{\updefault}NP}}}}}
\end{picture}
}
\caption{ 
The schematic structure, 
including the maximal cut, of the proof net for a relative
clause embedded within the matrix subject's complement.}
\label{f:scrc}
\end{figure}

\begin{figure}
\setlength{\unitlength}{0.00083333in}
\begingroup\makeatletter\ifx\SetFigFont\undefined%
\gdef\SetFigFont#1#2#3#4#5{%
  \reset@font\fontsize{#1}{#2pt}%
  \fontfamily{#3}\fontseries{#4}\fontshape{#5}%
  \selectfont}%
\fi\endgroup%
{\renewcommand{\dashlinestretch}{30}
\begin{picture}(7008,2463)(0,-10)
\path(2839,411)(2839,636)(3439,636)
\blacken\path(3379.000,606.000)(3439.000,636.000)(3379.000,666.000)(3379.000,606.000)
\path(3439,636)(3514,636)(3514,411)
\path(3664,411)(3664,936)(3139,936)
\blacken\path(3199.000,966.000)(3139.000,936.000)(3199.000,906.000)(3199.000,966.000)
\path(3139,936)(2239,936)(2239,411)
\path(1639,411)(1639,1236)(3439,1236)
\blacken\path(3379.000,1206.000)(3439.000,1236.000)(3379.000,1266.000)(3379.000,1206.000)
\path(3439,1236)(4639,1236)(4639,411)
\path(3139,1536)(1114,1536)(1114,411)
\path(4789,411)(4789,1536)(3139,1536)
\blacken\path(3199.000,1566.000)(3139.000,1536.000)(3199.000,1506.000)(3199.000,1566.000)
\dashline{60.000}(964,411)(964,1836)(3439,1836)
\blacken\path(3379.000,1806.000)(3439.000,1836.000)(3379.000,1866.000)(3379.000,1806.000)
\dashline{60.000}(3439,1836)(4939,1836)(4939,411)
\path(589,411)(589,2136)(3439,2136)
\blacken\path(3379.000,2106.000)(3439.000,2136.000)(3379.000,2166.000)(3379.000,2106.000)
\path(3439,2136)(5539,2136)(5539,411)
\texture{88555555 55000000 555555 55000000 555555 55000000 555555 55000000 
	555555 55000000 555555 55000000 555555 55000000 555555 55000000 
	555555 55000000 555555 55000000 555555 55000000 555555 55000000 
	555555 55000000 555555 55000000 555555 55000000 555555 55000000 }
\shade\path(3139,186)(64,186)(64,411)
	(3139,411)(3139,186)
\path(3139,186)(64,186)(64,411)
	(3139,411)(3139,186)
\path(5689,411)(5689,2436)
\blacken\path(5719.000,2376.000)(5689.000,2436.000)(5659.000,2376.000)(5719.000,2376.000)
\shade\path(3439,186)(6664,186)(6664,411)
	(3439,411)(3439,186)
\path(3439,186)(6664,186)(6664,411)
	(3439,411)(3439,186)
\put(3139,36){\makebox(0,0)[rb]{\smash{{{\SetFigFont{12}{14.4}{\familydefault}{\mddefault}{\updefault}the intern who the chance that the doctor}}}}}
\put(3439,36){\makebox(0,0)[lb]{\smash{{{\SetFigFont{12}{14.4}{\familydefault}{\mddefault}{\updefault}lost the reports bothered supervised the nurse}}}}}
\put(3289,711){\makebox(0,0)[b]{\smash{{{\SetFigFont{12}{14.4}{\familydefault}{\mddefault}{\updefault}NP}}}}}
\put(3289,1011){\makebox(0,0)[b]{\smash{{{\SetFigFont{12}{14.4}{\familydefault}{\mddefault}{\updefault}S}}}}}
\put(3289,1311){\makebox(0,0)[b]{\smash{{{\SetFigFont{12}{14.4}{\familydefault}{\mddefault}{\updefault}NP}}}}}
\put(3289,1911){\makebox(0,0)[b]{\smash{{{\SetFigFont{12}{14.4}{\familydefault}{\mddefault}{\updefault}WH}}}}}
\put(3289,2211){\makebox(0,0)[b]{\smash{{{\SetFigFont{12}{14.4}{\familydefault}{\mddefault}{\updefault}NP}}}}}
\put(3289,1611){\makebox(0,0)[b]{\smash{{{\SetFigFont{12}{14.4}{\familydefault}{\mddefault}{\updefault}S}}}}}
\end{picture}
}
\caption{ 
The schematic structure, 
including the maximal cut, of the proof net for a subject complement
clause embedded within a subject relative clause.}
\label{f:rcsc}
\end{figure}
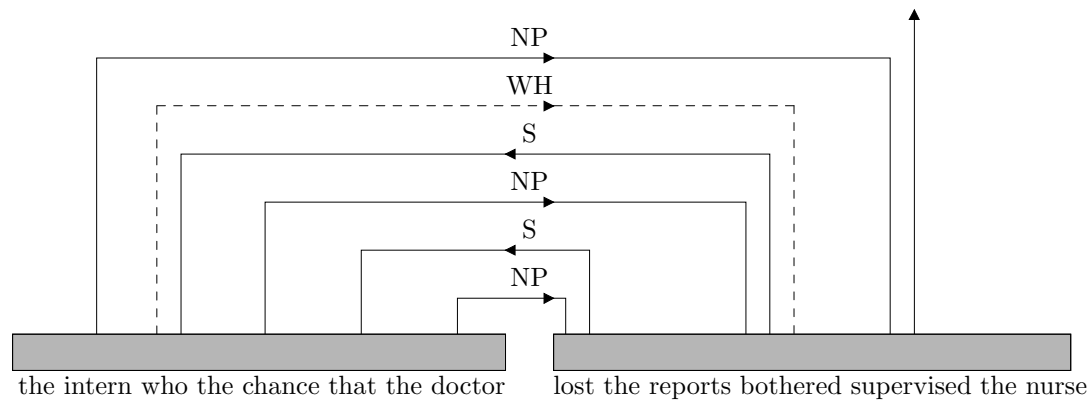

Thus far processing complexity seems proportional to the number of
axiom links crossed by a maximal cut.  The examples (\ref{e:gibson}),
reprinted here as (\ref{e:gibsonB}), 
suggest that the matter is more delicate.
Proof nets for these examples are schematically depicted in
Figures~\ref{f:scrc} and~\ref{f:rcsc} (only links crossed by the
maximal cut are shown).
\eenumsentence{ \label{e:gibsonB}
\item \label{e:scrcB}
The chance \const{SC}{ that the nurse \const{RC}{ who the doctor
supervised } lost the reports } bothered the intern.
\item \label{e:rcscB}
\bad The intern \const{RC}{ who the chance \const{SC}{ that the doctor 
lost the reports~} bothered } supervised the nurse. }

\namecite{Gibson96} present a hypothesis that accounts for
this difference in acceptability which is expressed in terms of the internal
states of an unspecified automaton that constructs a parse tree.  They
hypothesise that the human sentence processor overloads very quickly
when it predicts a category whose features are subsumed by the
features of one of the categories it is currently in the process of
completing.  In their approach a relative clause introduces a
prediction for an \S\ node with a \WH\ feature, while a sentential
complement introduces a prediction for an \S\ node without a \WH\
feature.  They propose that prediction for an \S\ node without a \WH\
feature is subsumed by a prediction for an \S\ node with a \WH\ feature,
so only (\ref{e:rcscB}) requires the introduction of a prediction
that is subsumed by the features of a category currently being
parsed.

This paper does not attempt to empirically evaluate this hypothesis,
but merely points out that in the context of the current examples,
this hypothesis can be restated as a constraint on the sequence of
axiom links crossed by any cut in the proof net.  To see this, note
that \namecite{Gibson96}'s category predictions correspond to
leftward-pointing axiom links in a proof net.  Thus there are two \S\
predictions at the locations of maximal cut complexity in the proof
nets in Figures~\ref{f:scrc} and~\ref{f:rcsc}.  

We could directly encode \namecite{Gibson96}'s proposal in a proof net
system by enriching the \S\ type with \WH\ features (possibly
inherited via some percolation mechanism, as they suggest): in proof
net terms their hypothesis would be that the human sentence processor
overloads if a leftward pointing axiom link is embedded within another
leftward pointing axiom link labelled with a subsuming type.

However, because \WH-dependencies appear in the proof nets used here
it is not necessary to enrich the \S\ type with \WH\ features, as the
relevant \S\ axiom links are easily identified geometrically in the
proof net (they are the ones introduced by expansion of a $\S / \WH$
type).  In the proof nets depicted in Figures~\ref{f:scrc}
and~\ref{f:rcsc} they are the \S\ axiom links adjacent to a \WH\ axiom
link.

\section{Incremental Semantic Interpretation}
The previous section has shown that theories of processing stated in
terms of incomplete dependencies can often be restated as geometric
constraints on proof nets.  Such constraints would be more plausible
if they can be shown to arise naturally from independently required
mechanisms or processes, such as incremental semantic interpretation.

As is well-known, one of the major attractions of resource logic
accounts of natural language is that the syntax-semantics interface
can take a particularly simple form, as the Curry-Howard
correspondence pairs each syntactic operation with a corresponding
semantic counterpart.

The idea explored in this section is basically as before: we divide
the utterance into two parts, and examine the complexity the
structural relationships between those parts.  In the previous
sections we examined a syntactic measure of complexity---the number of
incomplete dependencies---and argued that proof nets provide a
suitable structure for investigating these dependencies.  In this
section we investigate the complexity of the semantic interpretations
of the sequence of prefixes encountered as the sentence is heard and
processed.

Strong incremental interpretation requires that every prefix of a
sentence form a single semantic entity.  Take linguistic objects to be
pairs $X:\alpha$ of a type $X$ and a $\lambda$-term $\alpha$ of type
$X$, and let $\Gamma, \Delta$ be a string of lexical type/$\lambda$-term
pairs with an interpretation $\beta$, i.e.,
\[
        \seq{\Gamma, \Delta}{\S : \beta}
\]
Then an {\em incremental interpretation} of the prefix $\Gamma$
is an {\em interpolant} $X:\alpha$ that
satisfies the pair of constraints:
\[
 \seq{\Gamma}{X:\alpha} \qquad \mbox{and} \qquad
 \seq{X:\alpha, \Delta}{\S : \beta}.
\]
For many resource logics, including \LCG,
it is possible to prove that such incremental interpretations
always exist, i.e., if $\Gamma, \Delta$ has an interpretation
then $\Gamma$ has an incremental interpretation.

A natural question to ask is if there is any relationship
between the interpolants that function as incremental
interpretations and the proof net cuts used in the accounts
of center embedding processing complexity.

Unfortunately, the relationship between cut set size and the
complexity of the interpolant
is not straight-forward, because the interpolant
complexity can vary depending on the details of the logic.

Define the complexity of a type as one plus the number of binary
connectives appearing in it (thus $(\NP \backslash \S)/ \NP$ has
complexity 3).  Then the complexity of an interpolant
is never less than the number of axiom links crossed by the
corresponding cut, since the number of atomic links constructed
by expanding the interpolant using proof net connectives
is given by the interpolant's complexity.

On the other hand, the minimum complexity of an interpolant may be
greater than the number of atomic links crossed by a minimal
cut.  For example, consider the sequence of \LCG\ 
types corresponding to a transitive clause in a
{\small SOV} language.
\begin{equation} \label{e:sov}
  \NP:a, \; \NP:b, \, \marker \; \NP \backslash (\NP \backslash \S) : r
\end{equation}
The minimal cut at `\marker' in a proof net for (\ref{e:sov}) crosses
two \NP\ axiom links.  However, the smallest \LCG\ interpolant at
`\marker' is
\[
 \S / (\NP \backslash (\NP \backslash \S)) :
 \lambda f . f(b)(a),
\]
which has complexity 4.

Interestingly, this difference disappears if we enrich \LCG 's type with
product `$\ltimes$', with pair formation `$\tuple{\cdot,\cdot}$' as 
the corresponding semantic operation.  With this extension the
minimal interpolant is
\[
 \NP \ltimes \NP : \tuple{a,b}
\]
which corresponds directly to the axioms crossed by the minimal cut.

Thus the presence of an additional logical connective, in this
case `$\ltimes$', can alter the interpolant complexity, even if
that connective is not used in the grammar itself.  Thus interpolant
complexity depends on features of the logic used.

\section{Conclusion}
This paper has shown that proof nets provide a suitable formalization
of the notion of ``incomplete dependency'' used in many accounts of
processing complexity.  The structure of a proof net for a sentence
relevant for the complexity metrics discussed in this paper is usually
fixed by its meaning and standard linguistic assumptions about the
types of the words involved, so the results obtained do not depend
crucially on any particular linguistic theory.  More sophisticated
models of processing, such as \namecite{Gibson96}, appeal to the
arrangement of embedding relationships in a sentence.  The grapical
nature of proof nets makes it simple to express such theories in this
framework.

The last section of this paper investigated the relationship between
the incomplete dependencies expressed by proof nets and the complexity
of the types of incremental semantic interpretations.  However, it
seems that the complexity of the interpolants involved in incremental
interpretation depend crucially on the details of the logic used
to represent these interpolants.


\end{document}